\documentclass{article}

\usepackage{PRIMEarxiv}

\usepackage[utf8]{inputenc} 
\usepackage[T1]{fontenc}    
\usepackage{hyperref}       
\usepackage{url}            
\usepackage{booktabs}       
\usepackage{amsfonts}       
\usepackage{nicefrac}       
\usepackage{microtype}      
\usepackage{lipsum}
\usepackage{fancyhdr}       
\usepackage{graphicx}       
\usepackage{amsmath}
\usepackage{caption}
\usepackage{subcaption}
\graphicspath{{media/}}     

\title{Continually Learn to Map Visual Concepts to Large Language Models in Resource-constrained Environments}

\author{
    Clea Rebillard \\
     Bordeaux INP - ENSEIRB-MATMECA \\
    \texttt{clea.rebillard@gmail.com}
    \And
    Julio Hurtado \\
    CAMaCS, University of Warwick \\
    \texttt{julio.hurtado@warwick.ac.uk}\\
    \And
    Andrii Krutsylo \\
    Institute of Computer Science Polish Academy of Sciences \\
    \texttt{andrii.krutsylo@ipipan.waw.pl} \\
    \And 
    Lucia Passaro, Vincenzo Lomonaco \\
    University of Pisa \\
    \texttt{\{lucia.passaro, vincenzo.lomonaco\}@unipi.it} \\
    }

\begin{document}
\maketitle

\begin{abstract}
Learning continually from a stream of non-i.i.d. data is an open challenge in deep learning, even more so when working in resource-constrained environments such as embedded devices.
Visual models that are continually updated through supervised learning are often prone to overfitting, catastrophic forgetting, and biased representations. On the other hand, large language models contain knowledge about multiple concepts and their relations, which can foster a more robust, informed and coherent learning process. This work proposes Continual Visual Mapping (CVM), an approach that continually ground vision representations to a knowledge space extracted from a fixed Language model. Specifically, CVM continually trains a small and efficient visual model to map its representations into a conceptual space established by a fixed Large Language Model. Due to their smaller nature, CVM can be used when directly adapting large visual pre-trained models is unfeasible due to computational or data constraints. CVM overcome state-of-the-art continual learning methods on five benchmarks and offers a promising avenue for addressing generalization capabilities in continual learning, even in computationally constrained devices.
\end{abstract}


\section{Introduction}
Increasing the amount of training data has repeatedly shown significant positive impacts on the performance of deep learning models \cite{krizhevsky2012imagenet, openai2023gpt}. Generally speaking, as the quality and diversity of the training set expands, the performance of predictive models increases. This trend is particularly pronounced with \emph{Transformer} architectures \cite{vaswani2017attention}, which has uncovered new models capable of achieving state-of-the-art performance in multiple tasks, such as textual \cite{openai2023gpt}, visual \cite{dosovitskiy2020image} and multiple modalities \cite{radford2021learning}, provided that the necessary amount of data is available.

The Transformer architecture has shown remarkable performance, fostering the creation of a vast collection of large pre-trained models. These models can be used for multiple tasks, especially for individuals or entities lacking the necessary resources to train such large models. Since a considerable amount of data and computational power is required to fine-tune these models, much effort has been focused on enhancing their knowledge transferability \cite{lester2021power, hu2021lora}.

\begin{figure}[t]
    \centering
    \includegraphics[width=0.8\linewidth]{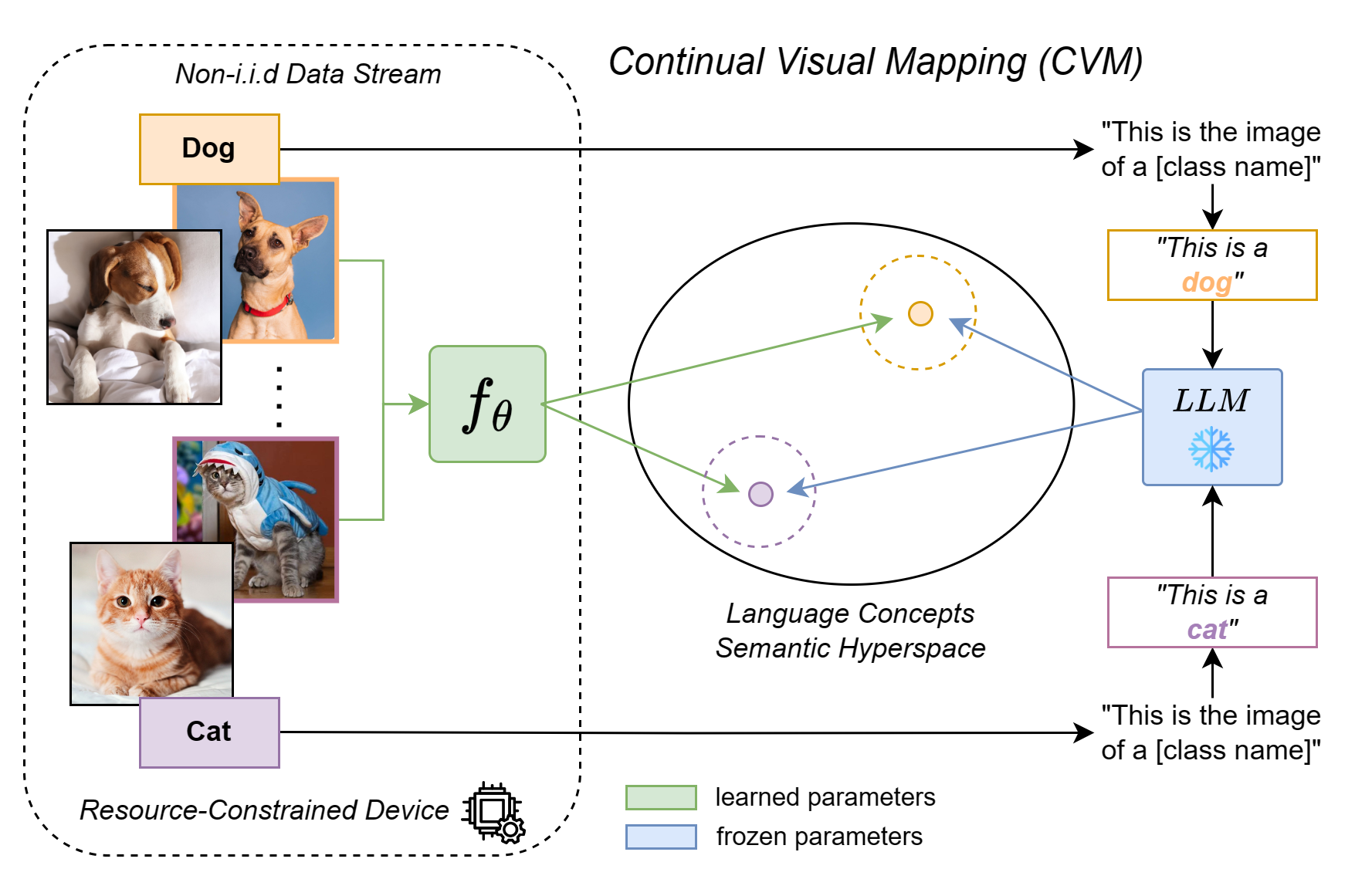}
    \caption{Conceptual image of our Continual Visual Mapping (CVM) method. A small and efficient parametric model $f_\theta$ is continually learned to map visual concepts into semantically richer embeddings seldomly extracted from frozen and pre-trained Large Language Models (LLMs).}
    \label{fig:method_init}
\end{figure}

Large pre-trained models have become popular in Continual Learning (CL). By fixing pre-trained weights and using their internal knowledge, many proposed methods have achieved state-of-the-art results in various settings \cite{wang2022dualprompt, wang2022learning, villa2023pivot}. However, continually adapting or fine-tuning large pre-trained models is computationally expensive and becomes impractical or even \emph{not possible} in resource-constrained environments. This issue extends to inference time since the time these models take to make a prediction is noticeably superior to smaller models.

The difficulty of using these large models in resource-constrained environments creates the following question: \emph{Can we still use knowledge extracted from large pre-trained models without using them directly in the continual learning and inference processes?}

We propose to use knowledge extracted from a \emph{frozen} Large Language Model (LLM) to train a smaller visual model, without using it directly in the training process. This strategy allows us to train and use this knowledge in resource-constrained environments. We decided to focus on language models for three main reasons: i) LLMs have shown to define inherently rich and robust semantic spaces that can be re-used across several tasks \cite{wei2022emergent}; ii) LLMs are rapidly evolving and generally more available (open-source) than visual or multi-modal models \cite{zhao2023survey}; iii) LLMs allow for a direct connection between the textual labels available in visual supervised datasets, and the concepts associated with them in language.

Hence, our approach focuses on training small visual models that can map their representations to a conceptual space established by a LLM, replacing the classifier with a distance function. Similar to CLIP \cite{radford2021learning}, we propose a multi-modal approach to train a visual model. However, instead of continually training the text and the visual transformer together, we obtain the representation from a frozen LLM and train only a small visual model to map the visual representations to this knowledgeable latent space. The visual model will learn the semantic information that lies in the fixed latent space. We call this approach \emph{Continual Visual Mapping (CVM)}, and a visual representation can be found in Figure \ref{fig:method_init}.

This approach is highly relevant in CL for two main reasons. Firstly, substituting the linear classifier with a stable set of anchor vectors eliminates interference generated in the classification layer, a recognized contributor to forgetting \cite{del2023studying, davari2022probing}. Secondly, we expect the model to learn more general visual representations that will transfer better to another task, which can increase forward transfer capabilities. 

Along with showing the performance of the proposed method in class-incremental and domain-incremental learning, we show that there are scenarios where even directly adapting large visual pre-trained models such as ViT can result in worse performance than CVM due to data scarcity or the specialized knowledge required. By training a smaller visual model, CVM also becomes significantly faster in inference time, making it practical for real-time applications. The main contributions of this work can be summarised as follows:

\begin{itemize}
    \item We propose a CL method that trains a small visual model by mapping its representations to knowledge space created by a frozen LLM (Sec. \ref{sec:cvm}). Improving performance and mitigating forgetting in standard benchmarks (Sec. \ref{sec:res}).
    \item We study the generalization and transfer capabilities of our proposal, showing favourable results against other CL methods (Sec. \ref{sec:for_trans}).
    \item We perform experiments in fine-grained datasets where a method based on a large pre-trained model fails. We show that CVM can achieve comparable performances (with lower inference time) with methods directly leveraging large models in training (Sec. \ref{sec:lim_pre-trained}). 
\end{itemize}

\section{Related Work}

\noindent\textbf{Continual Learning} Learning continually in deep learning \cite{parisi2019continual, delange2021continual, wang2023comprehensive} mainly aimed at addressing \emph{Catastrophic Forgetting} by enabling predictive models to learn new tasks without forgetting previously learned ones. Most CL methods can be categorized by different, non-exclusive computational approaches including: subdividing model parameters into subspaces for each new task \cite{rusu2016progressive, mallya2018piggyback}; imposing constraints on the learned gradients \cite{kirkpatrick2017overcoming, lopez2017gradient}; and using meta-learning to learn reusable weights for all tasks \cite{rajasegaran2020itaml, hurtado2021optimizing}. Out of these categories, memory-based methods such as Experience Replay (ER) \cite{rolnick2019experience, 10.1145/3147.3165} provide a straightforward solution that achieves good results. Memory-based methods address catastrophic forgetting by incorporating data from previous tasks into the training process for the current task \cite{ebrahimi2021remembering, buzzega2021rethinking}, and have been the state-of-the-art on most benchmarks since their ability to mitigate forgetting due to the constant repetition of previous tasks.

While using pre-trained models is not new in CL, the use of large pre-trained models has naturally shown better performance in several downstream tasks \cite{wang2022learning, jia2022visual, wang2022dualprompt, villa2023pivot}, since they can take advantage of the pre-trained representations that can significantly help the mitigation of catastrophic forgetting. These models are trained on vast amounts of data, allowing them to obtain valuable representations for different modalities and inputs \cite{dosovitskiy2020image, touvron2023llama}, making them very useful as a starting point for CL solutions. However, their massive amount of parameters and the need to access a considerable amount of data make them impractical to train (continually) as new distribution arrives \cite{cossu2022continual}.

\noindent\textbf{Learning Multi-Modal Representations} Training predictive models able to connect different data modalities (vision, text, speech, others) have been shown to provide richer representations and more reusable representations in deep learning \cite{openai2023gpt}. One of the most popular multi-modal models is CLIP \cite{radford2021learning}. Here, the authors train visual and language models to generate a latent space where visual and textual representations of the same input are close to each other. This fundamental insight has been further refined and expanded \cite{huang2023language, DBLP:journals/corr/abs-2102-05918, DBLP:journals/corr/abs-2006-06666} achieving state-of-the-art performance in multiple tasks \cite{DBLP:journals/corr/abs-2008-01392}.

Pioneering works \cite{NIPS2013_7cce53cf, weston2010large, socher2013zero} have explored training a visual model while using a static language representation (i.e., before the advent of contextualized embeddings and large language models). However, unlike previous works, we take advantage of extracting contextual information. Specifically, we build the knowledge space upon pre-trained LLM, expecting our visual model to benefit from the semantic knowledge contained within the LLM. Additionally, we explore how this approach can help reduce forgetting in CL scenarios by including a retention loss that mitigates the forgetting of the semantic knowledge of the previously learned tasks in the optimisation function. Such a solution provides a resource-efficient alternative to using large pre-trained models.

\section{Strategy}

\subsection{Background}
\noindent In supervised CL, we consider a stream of $T$ tasks. Each task $t$ consists of a new data distribution $D^t = (X^t, Y^t)$, where $X^t$ denotes the input instances and $Y^t$ denotes the instance labels. The goal is to train a classification model $ f_{\Theta}: X \longrightarrow Y $ using data from a sequence of $T$ tasks: $ D = \{D^1, ..., D^T \} $. During each task, model $f_{\Theta}$ will minimize the objective function $\mathcal{L}$ using data $D^t$.

\begin{equation}
    \mathcal{L}(D^t) = \frac{1}{N^t} \sum_{t=1}^{N^t} \mathcal{L}_t (f_{\Theta}(x_i^t), y_i^t)
    \label{eq:cl}
\end{equation}

Each task is presented sequentially to the model and trained for $E$ epochs. This work focuses on the traditional \emph{Class-Incremental (Class-IL)} setting. This scenario is often intended as the most challenging in CL \cite{van2019three}. Unlike \emph{Task Incremental Learning (Task-IL)}, in Class-IL, a task descriptor is only available during training. To verify the generality of our proposal, we also run experiments in a \emph{Domain Incremental (Domain-IL)} setting. In this last setting, each task considers all classes, but the distribution of each class changes over time.

In memory-based methods, together with minimizing Equation \ref{eq:cl}, model $f_{\Theta}$ needs to minimize $\mathcal{L}$ using the data available in the buffer memory $M$ at time $t$. The buffer $M^t$ comprises $|M|$ samples from previous distributions, meaning that at task $t$, the buffer will contain a limited set of samples collected from tasks $t'<t$. 

\begin{equation}
    \mathcal{L}_M(D^t, M^t) = \mathcal{L}(D^t) + \frac{1}{|M|} \sum_{i=1}^{|M|} \mathcal{L}_t (f_{\Theta}(m_i^t, y_i^t) 
    \label{eq:cl_memory}
\end{equation}

Instead of concatenating the dataset $D^t$ and $M^t$, which could neglect the representativeness of the memory, some works have adapted the concatenation of current data and memory at the batch level \cite{chaudhry2019tiny}. In this work, we follow this approach, which has been shown to increase the performance of memory-based methods.

\subsection{Continual Visual Mapping (CVM)}
\label{sec:cvm}

\begin{figure*}
\centering
  \begin{subfigure}{.5\textwidth}
    \includegraphics[width=\linewidth]{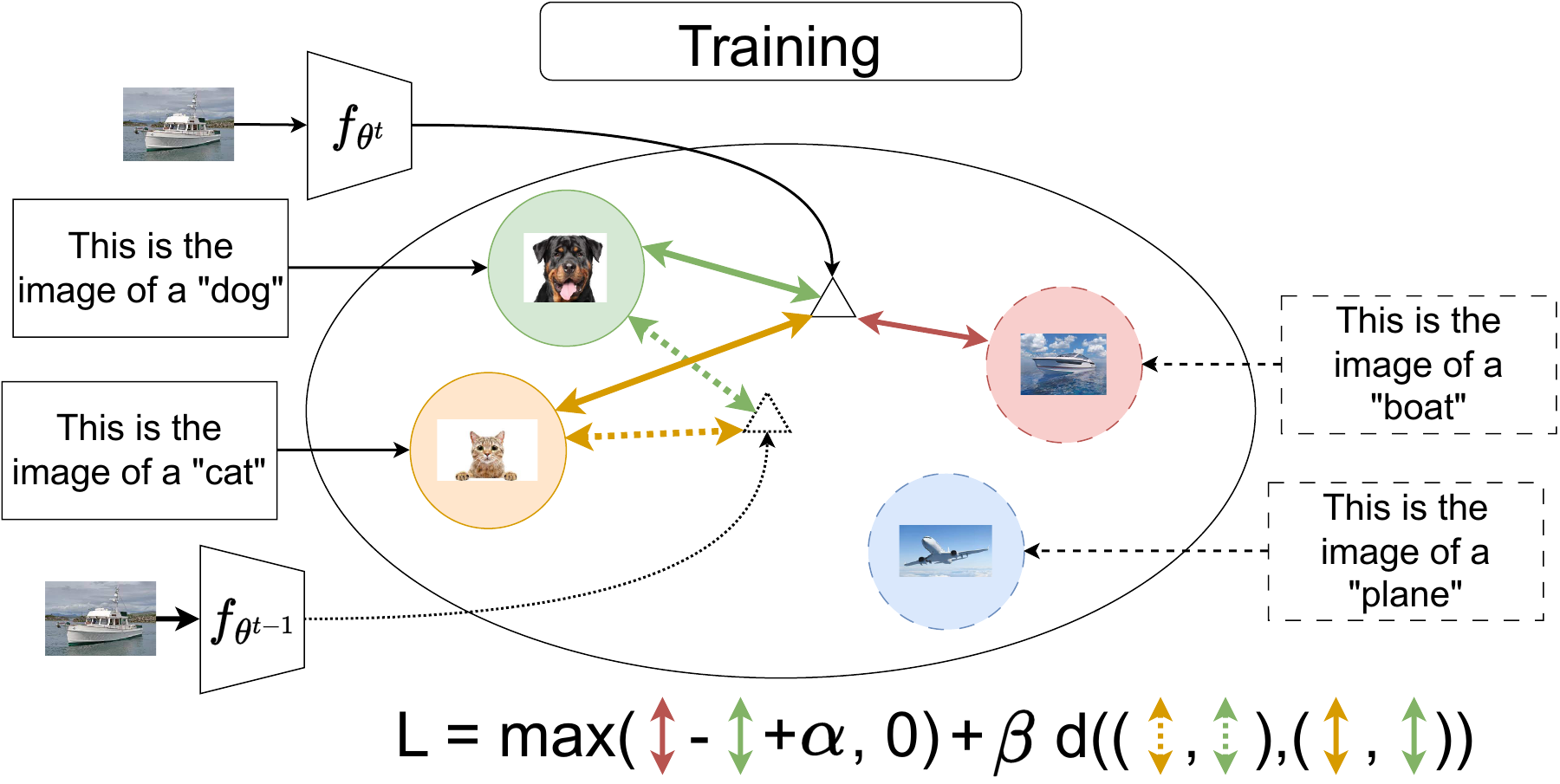}
    \caption{}
    \label{fig:cvm_train}
  \end{subfigure}%
  \hfill
  \begin{subfigure}{.5\textwidth}
    \includegraphics[width=\linewidth]{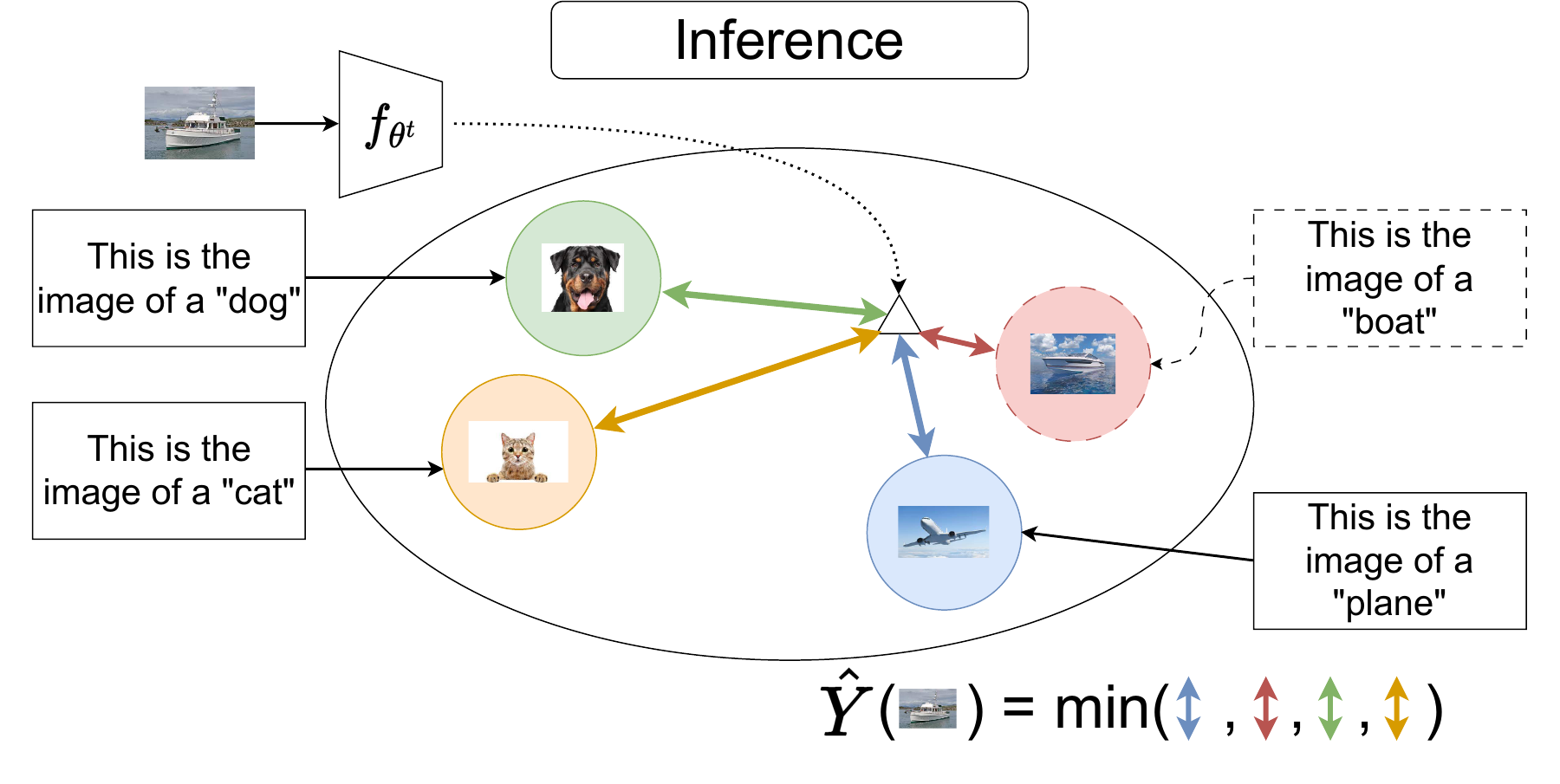}
    \caption{}
    \label{fig:cvm_inf}
  \end{subfigure}
\caption{Diagram of how CVM works at Training and Inference. During training (a), when adding new classes (plane and boat), two losses are involved: the triplet loss with a margin $\alpha$ (Equation \ref{eq:triplet}) and the Semantic distance loss restrained by coefficient $\beta$ (Equation \ref{eq:distill_loss}). At inference (b), the prediction $\hat{Y}$ is obtained by selecting the minimum distance between the image embedding and the seen classes in the latent space.}
\end{figure*}

\noindent In a classic formulation of a classification problem, a model $f_\Theta$ is composed of a feature extractor $f_\theta$ and a classifier $c_\omega$, been $\theta$ and $\omega$ the trainable weights of the model $f_\Theta$. During training, the learning strategy simultaneously tries to learn robust, semantic representations of the input ($f_\theta$) and a classifier ($c_\omega$) that uses these features to discriminate between the classes correctly, ideally generalising to samples not seen during training. 


In CL, the constant modification of the weights $\theta$ and $\omega$ results in ongoing interference between tasks. This interference can induce forgetting, which causes a decrease in the overall performance. We hypothesise that this interference is primarily due to the biases and boundaries the model learns during a task, knowledge that becomes unsuitable for future distributions. These biases are reinforced by the limited information the model receives through the input, which forces it to construct mappings based on limited visual information. Collectively, these factors strongly restrict the transferability of the learned representations to future tasks due to bias, which makes them prone to overfitting and less reusability.

Our proposed method replaces the classifier $c_\omega$ with a set of \emph{anchor vectors} representing the classes or concepts. These vectors are extracted from a pre-trained LLM, creating a fixed and knowledgeable latent space. The vectors establish a fixed latent space that the visual model can use as external knowledge. With this modification, we expect to (1) leverage the knowledgeable space created by the anchor vector to guide the training of the visual model and (2) increase the transferability of the visual representations.

Using a pre-trained LLM, we extract the current task $D^t$ representations, creating a latent space imbued with a broad understanding of the world \cite{manning2022human}. Given the absence of textual descriptions for images in most visual datasets, we generate a vector $C_c$ for each class $c$, using the prompt \emph{``This is an image of ''} + \textit{label}, where \textit{label} represents the text associated with each class (i.e., the class name). Importantly, the LLM is not employed during training, as our focus is solely on obtaining fixed vectors to replace the classifier. This strategy can be easily implemented in resource-constrained environments as a \emph{one-time query} to an external service.

Once the representations of classes $C$ at task $t$ are generated, we can obtain the corresponding class with a distance function. As shown in Equation \ref{eq:clss_min}, we compare the visual representations of the image $x_i$ with the known vectors in the latent space and select the nearest class. It is important to note that at time $t$, $C$ is composed only by seen classes. A visual representation of how the model works in inference can be found in Figure \ref{fig:cvm_inf}.

\begin{equation}
    \hat{Y_i} = min(d(f_\theta (x_i), C))
    \label{eq:clss_min}
\end{equation}

We use a Triplet Loss \cite{schroff2015facenet} to train the visual model. As shown in Equation \ref{eq:triplet}, the objective is to minimise the distance between the feature vector $f_\theta(x_i)$ and the representations of the corresponding class in the latent space $C_i^P$, while concurrently increasing the distance between $f_\theta(x_i)$ and negative classes. In all our experiments, we employ the cosine distance. The ultimate goal is to have the visual model map its representations to a fixed latent space created with knowledge gleaned from the LLM, thereby aligning both sources of information. This integration forms the basis of our approach, which we term Continual Visual Mapping (CVM).

\begin{equation}
\begin{split}
    L_m = \frac{1}{N} \sum_i^N max(0, & d(f_\theta(x_i), C_i^P) - d(f_\theta(x_i), C_i^N) + \alpha)
    \label{eq:triplet}
\end{split}
\end{equation}


In addition to mapping visual representations to the knowledge space, the visual model must grasp semantic information, specifically, the relationships between concepts. This knowledge acquisition is anticipated to enhance the generalization capabilities of the visual model, facilitated by the knowledge of the anchor vectors. However, the loss function $L_m$ operates under the assumption that all classes exhibit the same semantic distance, overlooking potential similarities between certain classes that share common characteristics and knowledge, such as the similarity between dogs and cats compared to aeroplanes.

To address this issue, we introduce a novel loss function designed to consider previous similarities learned by the model. Our proposal involves preserving the distances between classes encountered in the past. To achieve this, we calculate the distance between the representations of an input $x_i$ and its corresponding class $C^{t-1}$ using both the current model $f_{\theta^{t}}$ and the model at time $t-1$. As shown in Equation \ref{eq:distill_loss}, our objective is to minimize the discrepancy between these distances, thereby incorporating past similarities into the learning process.

\begin{equation}
\begin{split}
    L_d = \frac{1}{N} \sum_i^N d(&d(f_{\theta^{t}}(x_i), C^{t-1}), d(f_{\theta^{t-1}}(x_i), C^{t-1}))
    \label{eq:distill_loss}
\end{split}
\end{equation}

It is worth noting that Equation \ref{eq:distill_loss} only compares against classes seen before the current experiences to maintain the alignment in learned space.

We combine the losses according to Equation \ref{eq:loss_final}. We add a $\beta$ coefficient to control how much the model needs to remember previously learned distance between classes. Figure \ref{fig:cvm_train} visually represents the training loss.

\begin{equation}
    L_t = L_m + \beta L_d
    \label{eq:loss_final}
\end{equation}

\section{Experiments}

\noindent To showcase the effectiveness of CVM, we have empirically assessed it in various scenarios and benchmarks traditionally used in CL and following standard practices. We tested our method in a Class-IL scenario using CIFAR100 \cite{krizhevsky2009learning} and Tiny-ImageNet \cite{le2015tiny} datasets. We split these datasets into 10 experiences. We also run experiments in a Domain-IL scenario and higher-dimensional input spaces using the more realistic CORe50 dataset \cite{pmlr-v78-lomonaco17a}, designed explicitly for embedded continual object recognition applications. Here, the setting is composed of 10 domestic objects, and each new experience imposes a new class distribution (e.g., background, illumination, occlusion, scale, among others).

\subsection{Implementation Details}

\noindent We use a reduced version of a ResNet architecture proposed in \cite{rebuffi2017icarl} for the visual model to simulate a resource-constrained environment. The only difference between the model used in CVM and other baselines is that in CVM, we remove the classifier. This modification causes CVM to train fewer parameters. For the text model, we use SentenceBERT \cite{reimers2019sentence} to extract the textual representation for the classes. We ran all the experiments with three different seeds, each inducing a different ordering of the sequences of tasks. 

We compare our proposal to different types of methods. First, we compare our method against some classic regularization-based CL methods like AGEM \cite{chaudhry2018efficient}, EWC \cite{kirkpatrick2017overcoming}, and LwF \cite{li2017learning}. We also compare against memory-based methods like ER \cite{chaudhry2019tiny}, DER++ \cite{buzzega2020dark} and iCarl \cite{rebuffi2017icarl}. iCarl is especially interesting to our case since it also removes the classifier at inference time. However, the methods differ mainly in training, as iCarl uses a cross-entropy loss with labels instead of the distance to knowledge anchor vectors. The way iCarl generates the vector per class is also different since the vectors are generated based on samples in the memory.

In all experiments, we use an SGD optimizer with a learning rate of $0.1$ and a batch size of 32. CIFAR100 and TinyImagenet train for 50 epochs per experience; for CORe50, we train for 30 epochs. We use the implementation provided by Avalanche \cite{lomonaco2021avalanche}. We used the hyper-parameters proposed by the authors in the corresponding paper. The code of CVM will become available upon acceptance.

We evaluate the average accuracy (Acc) and forgetting (For) over the $T$ tasks after the sequential learning, proposed in \cite{lopez2017gradient}. Equations \ref{eq:metrics_acc} and \ref{eq:metrics_for} show the formula for the accuracy and forgetting, respectively, where $Acc_{i,j}$ is the accuracy of task $i$ after training task $j$.

\begin{equation}
Acc = \frac{1}{T} \sum_{i=1}^T Acc_{T,i}
\label{eq:metrics_acc}
\end{equation}

\begin{equation}
For = \frac{1}{T-1} \sum_{i=1}^{T-1} Acc_{T,i} - Acc_{i,i}
\label{eq:metrics_for}
\end{equation}

\subsection{Results}
\label{sec:res}

\begin{table}[]
\begin{center}
\caption{Accuracy and forgetting for CIFAR100, Tiny-ImageNet, and CORe50. CVM outperforms all methods regarding average accuracy due to the combination of $L_m$ and the semantic loss. It is worth noting that we cannot measure the forgetting metric in CORe50 since it is designed with a unique and fixed test set.}
\setlength{\tabcolsep}{3pt}
\begin{tabular}{l|cc|cc|c}
\toprule
 & \multicolumn{4}{c|}{Class-IL} & Domain-IL \\
 & \multicolumn{2}{c|}{CIFAR100} & \multicolumn{2}{c|}{Tiny-ImageNet} & CORe50 \\
 & Acc & For & Acc & For & Acc \\ 
 \hline
 Naive     & 9.0\% \begin{tiny}$\pm 1.7$\end{tiny} & 78.6\% \begin{tiny}$\pm 0.8$\end{tiny} 
           & 7.4\% \begin{tiny}$\pm 0.1$\end{tiny} & 73.9\% \begin{tiny}$\pm 0.3$\end{tiny} 
           & 23.1\% \begin{tiny}$\pm 1.0$\end{tiny} \\
 AGEM      & 9.3\% \begin{tiny}$\pm 0.9$\end{tiny} & 77.8\% \begin{tiny}$\pm 0.3$\end{tiny}
           & 7.5\% \begin{tiny}$\pm 0.1$\end{tiny} & 74.7\% \begin{tiny}$\pm 0.3$\end{tiny} 
           & 26.0\% \begin{tiny}$\pm 1.3$\end{tiny} \\
 LWF       & 8.8\% \begin{tiny}$\pm 1.5$\end{tiny} & 76.3\% \begin{tiny}$\pm 1.0$\end{tiny}
           & 7.4\% \begin{tiny}$\pm 0.0$\end{tiny} & 73.0\% \begin{tiny}$\pm 0.4$\end{tiny} 
           & 31.7\% \begin{tiny}$\pm 2.2$\end{tiny}  \\
 EWC       & 8.6\% \begin{tiny}$\pm 0.2$\end{tiny} & 77.7\% \begin{tiny}$\pm 0.7$\end{tiny}
           & 7.3\% \begin{tiny}$\pm 0.1$\end{tiny} & 70.7\% \begin{tiny}$\pm 0.6$\end{tiny} 
           & 24.0\% \begin{tiny}$\pm 3.4$\end{tiny} \\
 \hline
 ER        & 21.0\% \begin{tiny}$\pm 0.1$\end{tiny} & 65.0\% \begin{tiny}$\pm 0.7$\end{tiny}
           & 13.1\% \begin{tiny}$\pm 0.2$\end{tiny} & 65.8\% \begin{tiny}$\pm 0.2$\end{tiny} 
           & 33.0\% \begin{tiny}$\pm 0.4$\end{tiny} \\
 + $L_m$   & 20.9\% \begin{tiny}$\pm 0.7$\end{tiny} & 64.7\% \begin{tiny}$\pm 0.5$\end{tiny}
           & 10.7\% \begin{tiny}$\pm 0.3$\end{tiny} & 66.2\% \begin{tiny}$\pm 0.6$\end{tiny} 
           & 32.8\% \begin{tiny}$\pm 1.3$\end{tiny} \\
 \hline
 ER + EWC  & 19.2\% \begin{tiny}$\pm 0.4$\end{tiny} & 64.8\% \begin{tiny}$\pm 0.7$\end{tiny}
           & 17.7\% \begin{tiny}$\pm 1.0$\end{tiny} & 55.5\% \begin{tiny}$\pm 0.6$\end{tiny} 
           & 31.9\% \begin{tiny}$\pm 0.1$\end{tiny} \\
 + $L_m$   & 25.2\% \begin{tiny}$\pm 0.6$\end{tiny} & 51.9\% \begin{tiny}$\pm 0.2$\end{tiny}
           & 11.5\% \begin{tiny}$\pm 1.2$\end{tiny} & 58.7\% \begin{tiny}$\pm 0.8$\end{tiny} 
           & 32.2\% \begin{tiny}$\pm 2.3$\end{tiny} \\
 \hline
 DER       & 19.1\% \begin{tiny}$\pm 0.8$\end{tiny} & 65.3\% \begin{tiny}$\pm 0.1$\end{tiny}
           & 3.9\% \begin{tiny}$\pm 0.4$\end{tiny} & 64.1\% \begin{tiny}$\pm 0.7$\end{tiny} 
           & 34.8\% \begin{tiny}$\pm 2.3$\end{tiny} \\
 + $L_m$   & 21.0\% \begin{tiny}$\pm 1.2$\end{tiny} & 64.4\% \begin{tiny}$\pm 0.6$\end{tiny}
           & 11.7\% \begin{tiny}$\pm 0.4$\end{tiny} & 64.9\% \begin{tiny}$\pm 0.5$\end{tiny} 
           & 33.1\% \begin{tiny}$\pm 0.6$\end{tiny} \\ 
 \hline
 iCarl     & 29.1\% \begin{tiny}$\pm 1.1$\end{tiny} & 23.2\% \begin{tiny}$\pm 0.2$\end{tiny}
           & 24.3\% \begin{tiny}$\pm 0.2$\end{tiny} & 16.2\% \begin{tiny}$\pm 0.1$\end{tiny} 
           & 17.8\% \begin{tiny}$\pm 1.8$\end{tiny} \\
 CVM       & \textbf{32.9\%} \begin{tiny}$\pm 0.4$\end{tiny} & 40.1\% \begin{tiny}$\pm 0.8$\end{tiny}
           & \textbf{27.0\%} \begin{tiny}$\pm 0.8$\end{tiny} & 33.9\% \begin{tiny}$\pm 0.8$\end{tiny} 
           & \textbf{37.6\%} \begin{tiny}$\pm 1.1$\end{tiny} \\
 \bottomrule
\end{tabular}
\label{tab:cl_res_ci}
\end{center}
\end{table}

\noindent The results for Class-IL benchmarks are provided in Table \ref{tab:cl_res_ci}. As previous studies have shown, regularization methods are suboptimal in this scenario. However, performance gains can be achieved by leveraging access to previous tasks stored in a buffer, exemplified by ER and DER++. Additionally, combining EWC with classical ER only seems to increase performance in Tiny-ImageNet.

To boost the efficacy of memory-based approaches, replacing Cross-Entropy Loss with the mapping loss (Equation \ref{eq:triplet}) proves to be a powerful learning strategy. This strategy significantly increases accuracy, particularly in ER + EWC, where the accuracy increases $6\%$. The model gains valuable insights by incorporating additional information from latent representations, which is particularly advantageous in CL scenarios where knowledgeable representations can enhance generalization. However, it is noteworthy that this transferability is most pronounced in simple benchmarks like CIFAR100. In contrast, on more complex benchmarks such as Tiny-ImageNet and CORe50, $L_m$ does not contribute to performance improvement. This is primarily attributed to higher forgetting, indicating that complex benchmarks need extra encouragement to mitigate forgetting.

To mitigate the forgetting, we propose the novel loss $L_d$. This loss is specifically crafted to influence the distances between classes, acknowledging that certain classes are more similar than others, which is not considered in the traditional triplet loss. Combining this loss with $L_m$ creates CVM, which aims to extract valuable knowledge from the latent space while preserving the inherent class relationships. Our experimental results, as detailed in Table \ref{tab:cl_res_ci}, confirm that CVM outperforms previous methods, including iCarl, which also removes the classifier during inference. These findings highlight that \textit{relying solely on information extraction from an LLM is insufficient}, emphasizing the necessity of incorporating additional mechanisms in CL scenarios.

Similar results can be seen in CORe50, as shown in Table \ref{tab:cl_res_ci}. Most regularization techniques find it challenging to perform reasonably, and memory-based methods produce the best results. iCarl, despite being a memory-based method, also faces difficulties due to the constant shift of the class distribution. Here, introducing $L_m$ does not aid ER nor DER++ in enhancing their outcomes. However, CVM still surpasses its predecessors by encouraging the reuse of previous distances.

\subsection{Ablation}

\begin{table}[]
\begin{center}
\caption{We use the CIFAR100 benchmark to test the performance with different memory sizes. Methods scale differently; iCarl does not scale well, increasing only $10\%$, DER increases performance quite a bit ($31\%$). CVM still outperform in every scenario.}
\setlength{\tabcolsep}{3pt}
\begin{tabular}{l|ccccc}
 & \multicolumn{5}{c}{Memory Size} \\
 & 500 & 1000 & 2000 & 3000 & 5000 \\
 \hline
 ER       & 21.0\% & 28.2\% & 36.8\% & 40.5\% & 47.3\% \\
 + $L_m$    & 20.9\% & 27.9\% & 37.7\% & 41.8\% & 47.3\%  \\
 \hline
 ER + EWC & 19.2\% & 26.1\% & 35.0\% & 39.4\% & 43.6\% \\
 + $L_m$    & 25.2\% & 26.1\% & 35.5\% & 39.1\% & 46.8\%  \\
 \hline
 DER      & 19.1\% & 26.6\% & 39.3\% & 45.3\% & 50.9\% \\
 + $L_m$    & 21.0\% & 28.2\% & 39.6\% & 43.1\% & 47.5\% \\
\hline
iCarl     & 29.1\% & 36.4\% & 37.7\% & 38.8\% & 39.6\% \\
CVM       & \textbf{32.9\%} & \textbf{38.7\%} & \textbf{43.5\%} & \textbf{47.5\%} & \textbf{51.4\%} \\
\end{tabular}
\label{tab:cl_res_mem}
\end{center}
\end{table}

\noindent In CL, preserving the plasticity of the model is fundamental to learning new tasks and distributions. However, the trade-off between learning new classes and forgetting the previous one is the primary concern that most CL methods face.

In memory-based methods, increasing the memory size improves the representativeness of the buffer, which should increase performance by decreasing forgetting. As shown in Table \ref{tab:cl_res_mem}, DER++ is the method that scales the best as we increase the buffer size, significantly increasing its performance; however, it still achieves worse results than CVM, which consistently outperforms previous methods. On the other hand, iCarl scales poorly.

Something that explains the low performance of iCarl is plasticity. As shown in Table \ref{tab:cl_res_ci}, iCarl obtained a lower forgetting value. These results can be explained by the significant restrictions that iCarl imposes when training new tasks. These constraints become more evident as we increase the memory size. iCarl can maintain a lower level of forgetfulness but at the cost of having low plasticity and being unable to scale up when we increase memory size. In contrast, CVM has no problem scaling up and maintaining a high level of plasticity. 

We present two compelling arguments affirming the effectiveness of CVM and the importance of the proposed learning strategy for CL. Firstly, our approach involves mapping visual representations to a stable and well-informed space, removing the need for a classifier. Previous studies highlight the classifier as a critical contributor to model overfitting, particularly in challenging CL environments \cite{del2023studying}. This finding gains further support from the insights in Table \ref{tab:cl_res_mem}, where enhancing memory representativeness enhances overall performance and mitigates the risk of overfitting to limited memory capacities. Secondly, the strategic alignment of concepts in the latent space is crucial. The semantic information within the space generated by a frozen LLM discerns the similarities among diverse classes. This concept is actively promoted by incorporating the loss described in Equation \ref{eq:distill_loss}.

\begin{figure}[tb]
  \centering
  \begin{subfigure}{0.48\linewidth}
    \includegraphics[width=0.98\linewidth]{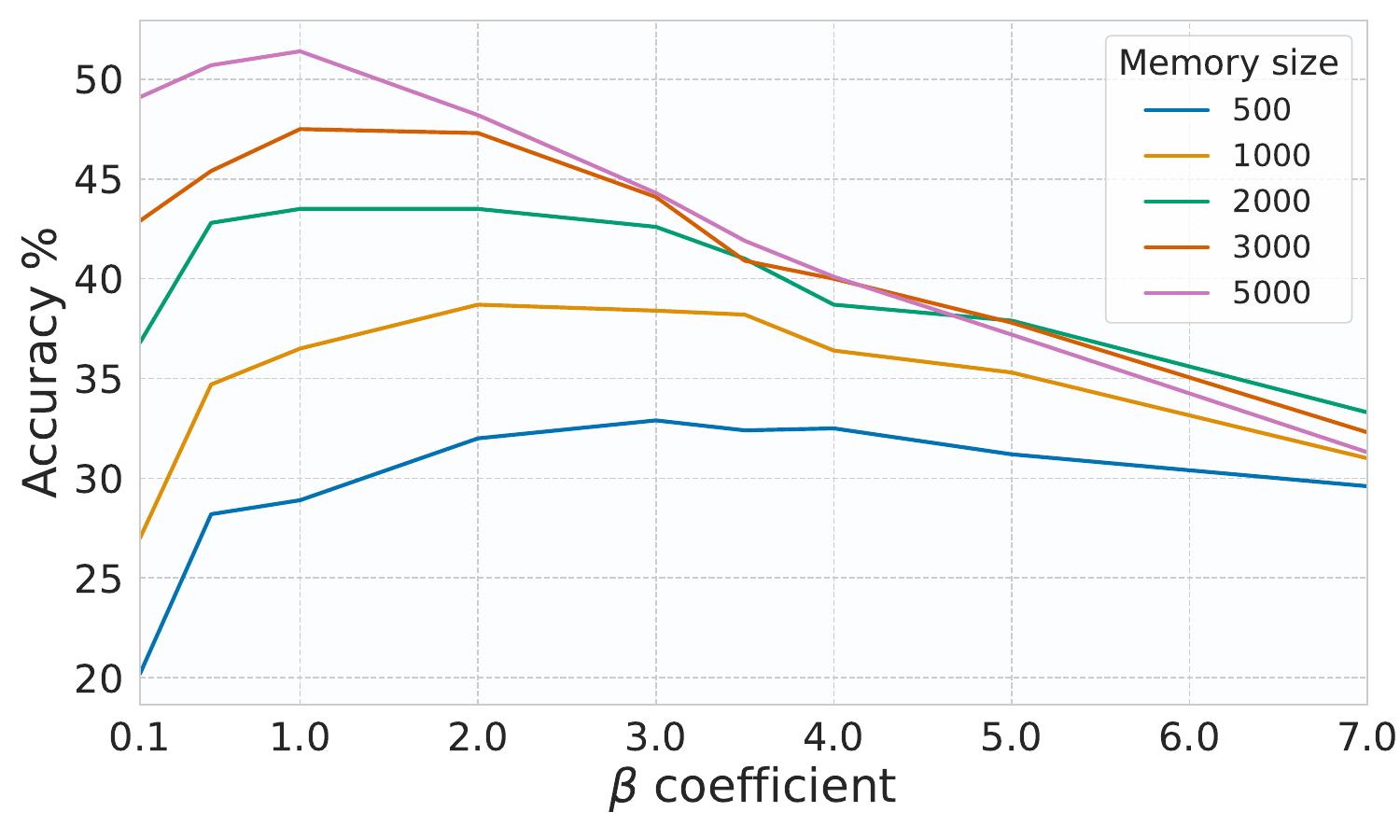}
    \caption{}
    \label{fig:acc_beta}
  \end{subfigure}
  \hfill
  \begin{subfigure}{0.48\linewidth}
    \includegraphics[width=0.98\linewidth]{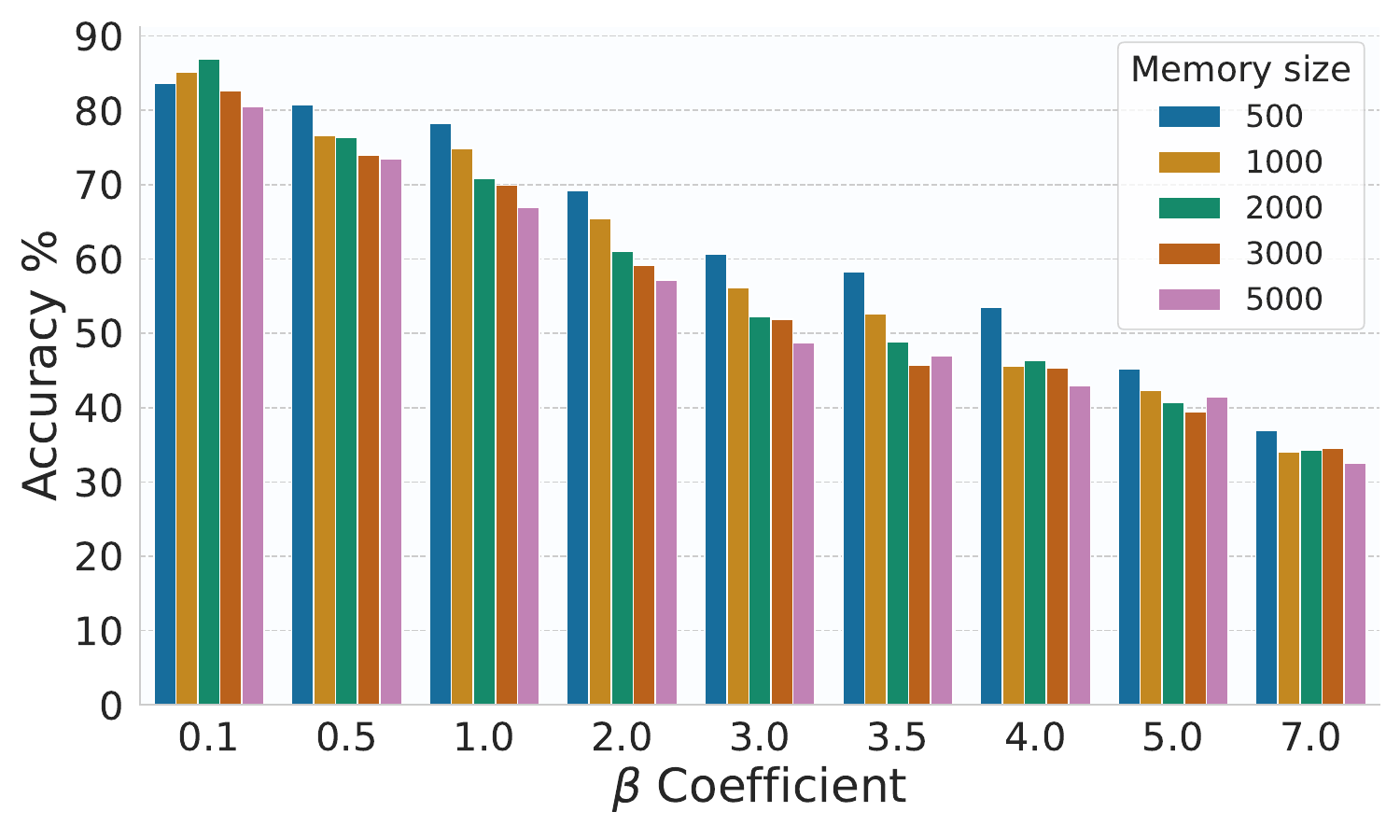}
    \caption{}
    \label{fig:acc_last_task}
  \end{subfigure}
  \caption{Ablation study over the $\beta$ coefficient. Intuitively, $\beta$ controls how much to consider past distances, impacting performance. In \ref{fig:acc_beta}, it is shown that the optimal coefficient tends to decrease as we increase the memory size since memory provides a better representation of past distributions. On the other side, $\beta$ controls the plasticity of a model. Figure \ref{fig:acc_last_task} shows the accuracy of the last task. Increasing $\beta$ encourages the model to maintain previous semantic knowledge while decreasing plasticity.}
\end{figure}

\noindent \textbf{CVM components} A critical component of CVM is the $\beta$ coefficient, which, as shown in Figure \ref{fig:acc_beta}, the optimal value is strongly related to the memory size. As we increase the memory size, the model can repeat more concepts presented in the buffer, which decreases the need to remember the exact distance of previous tasks, meaning that CVM can work with a smaller $\beta$. On the other hand, if we decrease the memory size, the model must depend mainly on the loss function to maintain semantic knowledge learned in the past.

The second experiment about $\beta$ concerns the plasticity of the model. As shown in figure \ref{fig:acc_last_task}, the accuracy of the last task will decrease as we increase the $\beta$ coefficient. This is explained by discouragement in moving the weights when training the new task, reducing the plasticity. With a high value of $\beta$, the loss promotes fewer modifications on the model weights, reducing forgetting but decreasing the final accuracy.

Another critical component of CVM is the LLM from where to create the fixed latent space. Maintaining the prompt fixed, we compare SentenceBERT with the textual model of CLIP and BERT. Results are shown in Table \ref{tab:text_model}. These results show that using an appropriate model to generate the latent space is essential. However, future work will explore using more detailed descriptions or concepts extracted from the image instead of directly using the class.

\begin{table}
    \centering
    \caption{The knowledgeable space created by the LLM is critical for CVM. As the performance of these models grows, our method should obtain a better approximation of a semantic space. Here, we test different LLMs from where to extract the world representation.}
    \setlength{\tabcolsep}{3pt}
    \begin{tabular}{c|ccc}
             & SentenceBERT  & Text CLIP   & BERT \\
         \hline
         Acc & 32.9\%\begin{tiny}$\pm 0.4$\end{tiny} & 31.0\%\begin{tiny}$\pm 0.2$\end{tiny} & 29.8\%\begin{tiny}$\pm 0.6$\end{tiny}\\
    \end{tabular}
    \label{tab:text_model}
\end{table}

\section{Generalization Capabilities}
\label{sec:for_trans}

\noindent For a continual learner to be efficient, it should reuse previous knowledge when learning new tasks. It is hypothesised that reusing previous knowledge could mitigate forgetting since the model would have no incentive to modify that knowledge. On this note, an essential feature in CL models is the generalization ability due to the transfer capability to \emph{future tasks}.

There are two distinct approaches to assessing the transfer capability of a model. The first approach involves freezing the model and training a linear classifier (linear probing) on unseen classes. This evaluation method measures the generality of the representations acquired by the model for future tasks. The second approach focuses on evaluating the zero-shot capabilities of the model. In the case of CVM, this involves incorporating an anchor vector of unseen classes and verifying the accuracy exclusively for those classes while keeping the visual model frozen. This method provides insights into the model's ability to classify classes that are not part of the training set.

\begin{figure}[tb]
  \centering
  \begin{subfigure}{0.48\linewidth}
    \includegraphics[width=\linewidth]{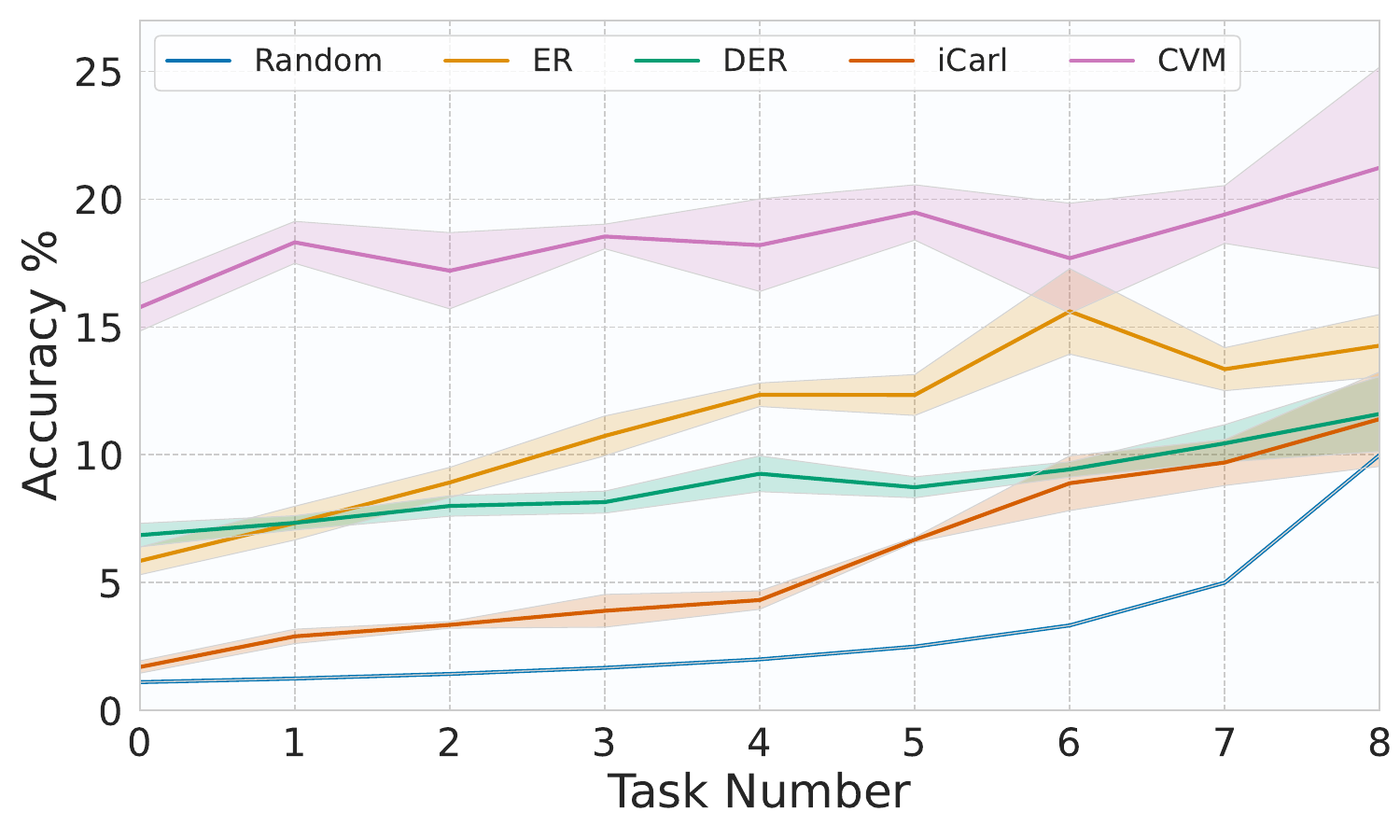}
    \caption{}
    \label{fig:ford_transfer}
  \end{subfigure}
  \hfill
\begin{subfigure}{0.48\linewidth}
    \includegraphics[width=\linewidth]{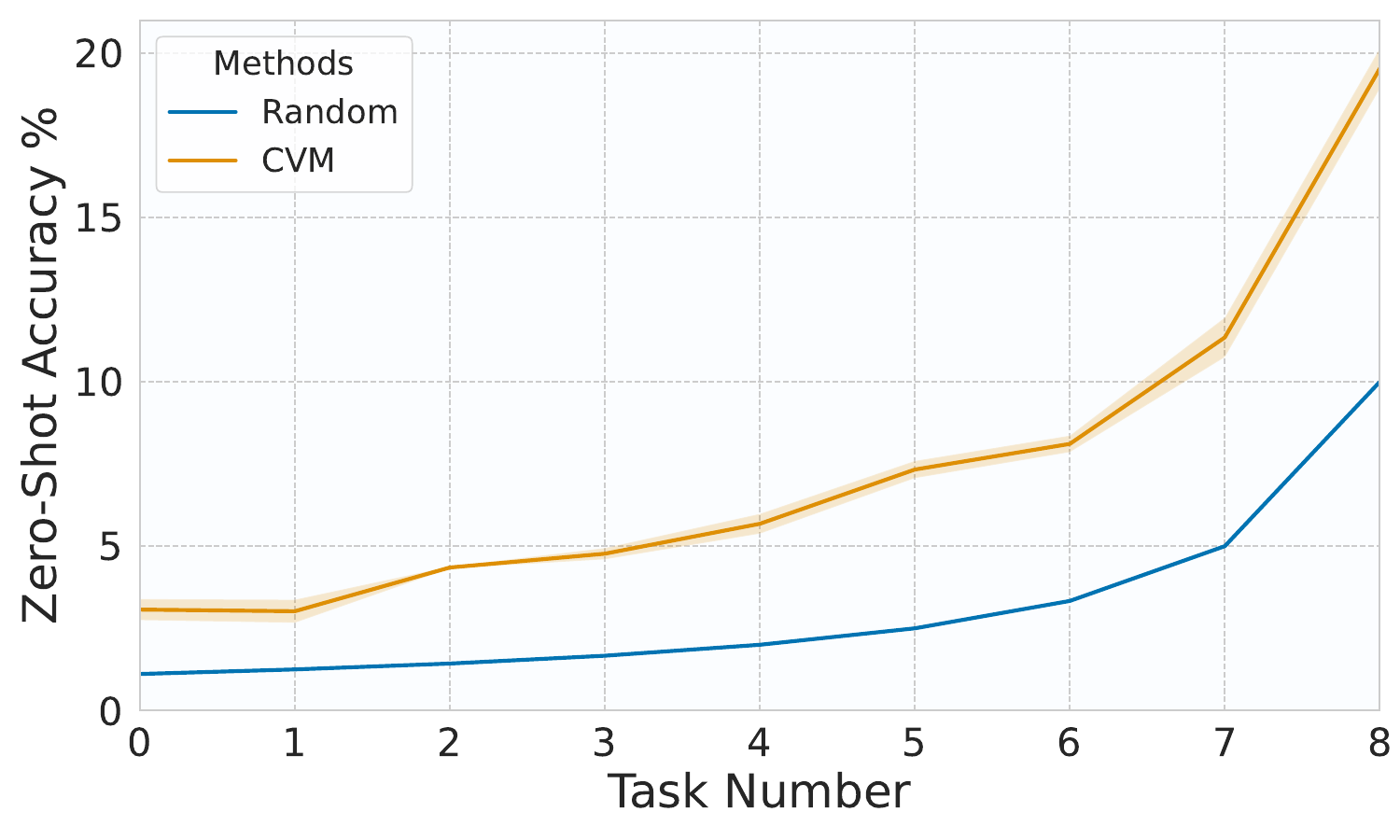}
    \caption{}
    \label{fig:zero_shot_cap}
  \end{subfigure}
  \caption{Using CIFAR100, we evaluate the approaches over unseen tasks, meaning that after Task 0, we have 90 unseen classes. Figure \ref{fig:ford_transfer} shows the accuracy when freezing the model and learning a linear classifier with the unseen classes after training each task. Random denotes a random assignation of classes. Figure \ref{fig:zero_shot_cap} shows the accuracy of the zero-shot performance of CVM compared to a random classification. After training task $t$, we test the zero-shot capabilities on classes of tasks $t' > t$ and compare them with randomly assigned classes: $\nicefrac{1}{|unseen|}$.}
\end{figure}

In the first approach, reusable representations indicate how prepared a model is to learn new tasks. A low value in accuracy indicates that the model should highly modify its representation to learn new tasks; on the other hand, a higher accuracy value indicates that minor modifications are needed to adjust to new tasks. The results of applying the linear probing are shown in Figure \ref{fig:ford_transfer}. The results indicate that the learned representations of CVM have the highest transfer capabilities, ranging from 16\% when comparing with 90 unseen classes (after training the first task) to about 22\% when only 10 classes remain. In contrast, iCarl fails to generate good representations, as it has close to random values.

The zero-shot capabilities take the previous observation to an extreme. If a model can learn to identify a class without modifying its weights, it learns the new distribution without forgetting. However, methods with a classifier are limited to the classes seen due to constraints in how the classifier is optimized and structured. In contrast, CVM holds a distinct advantage. By relying on a distance metric instead of a trained classifier, CVM can classify classes not encountered during training by adding an anchor vector that represents the textual description of the new class. In this context, the zero-shot accuracy performance of the model is computed on unseen classes for tasks $t' > t$ when $t$ is the current task. Results are illustrated in Figure \ref{fig:zero_shot_cap}. It is noteworthy that methods with a classifier exhibit a performance of $0\%$ due to the inherent limitations of the classifier.

Unlike accuracy, where the final value represents how well methods perform, the transfer capabilities are better measured over the complete sequence. We developed a simple scoring system to compare the transfer capabilities between various methods easily. This score is obtained by averaging the accuracy achieved by each metric after completing each task. For instance, in the case of linear probing, we train a linear classifier ($f_\gamma^t$) after each task ($t$), using only data from unseen tasks ($t'>t$), i.e., $X^{t'>t}$ and $Y^{t'>t}$. This score helps us understand how well the method performs on unseen classes. We average all these performances, as shown in Equation \ref{eq:fw_score}.

\begin{equation}
    FW_{score} = \frac{1}{T - 1} \sum_{t=1}^{T-1} Acc(f_\gamma^t, X^{t'>t}, Y^{t'>t})
    \label{eq:fw_score}
\end{equation}

A higher value means that the representations learnt during the sequence are more capable of being used in future tasks. The results of the score for each method can be found in Table \ref{tab:fw_score}.

\begin{table}[]
\begin{center}
\caption{The $FW_{Score}$ is a metric that helps compare the transfer capacity of different CL methods. CVM outperforms previous methods of learning representations that learn unseen classes using two approaches.}
\setlength{\tabcolsep}{3pt}
\begin{tabular}{l|ccccc}
& Random & ER & DER & iCarl & CVM \\
\hline
$FW_{Score}$  & 3.1\% & 11.2\% & 8.9\% & 5.9\% & 18.4\% \\
Zero-shot Acc.  & 3.1\% & - & - & - & 7.9\% \\
\end{tabular}

\label{tab:fw_score}
\end{center}
\end{table}

\section{Limitations of Pre-trained Models}
\label{sec:lim_pre-trained}

\noindent Even in scenarios where we have ample computational power, other issues can make methods based on large visual pre-trained models unsuitable, e.g., the need for more data, the specificity or even the granularity of the problem. 

We introduce two Continual Learning (CL) benchmarks utilizing the GTSRB \cite{Houben-IJCNN-2013} and Aircraft \cite{maji13fine-grained} datasets. For the GTSRB dataset, we delineate 10 tasks, while the Aircraft dataset is partitioned into 5 tasks, both adhering to a Class-IL scenario.

The primary challenge confronted by methods relying on pre-trained models arises from the granularity of the dataset. This fine-grained nature necessitates additional information not present in the weights of large pre-trained models. Consequently, pre-trained models require assistance to learn the sequence, leading to low performance. In Table \ref{tab:cl_res_clip}, we present the performance of L2P \cite{wang2022learning} both with and without memory. Notably, these approaches fail to outperform methods employing simpler models. Although pre-trained models may perform similarly to simpler and smaller models, their longer inference times can adversely impact overall performance.

\begin{table}[]
\begin{center}
\caption{Two CL benchmarks were created with granular datasets. In these benchmarks, visual models outperform CL methods based on large pre-trained models in terms of accuracy and inference time.}
\setlength{\tabcolsep}{3pt}
\begin{tabular}{l|cc|cc}
& \multicolumn{2}{c|}{GTSRB} & \multicolumn{2}{c}{Aircraft} \\
& Acc & Inf. Time & Acc & Inf. Time \\
\hline
L2P      & 32.4\% \begin{tiny}$\pm 0.4$\end{tiny} & $\sim$ 188    & 34.0\% \begin{tiny}$\pm 3.7$\end{tiny} & $\sim$ 100 \\
L2P + ER & 74.9\% \begin{tiny}$\pm 4.1$\end{tiny} & $\sim$ 188    & 34.9\% \begin{tiny}$\pm 1.3$\end{tiny} & $\sim$ 100 \\
ER       & 79.4\% \begin{tiny}$\pm 0.2$\end{tiny} & $\sim$ 14     & 35.0\% \begin{tiny}$\pm 1.6$\end{tiny} & $\sim$ 65 \\
DER      & 83.6\% \begin{tiny}$\pm 1.4$\end{tiny} & $\sim$ 14     & 36.1\% \begin{tiny}$\pm 2.3$\end{tiny} & $\sim$ 65 \\
CVM      & \textbf{84.8\%} \begin{tiny}$\pm 1.3$\end{tiny} & $\sim$ 14 & \textbf{36.5\%} \begin{tiny}$\pm 1.1$\end{tiny} & $\sim$ 65 \\  
\end{tabular}
\label{tab:cl_res_clip}
\end{center}

\end{table}

\section{Conclusions}

\noindent We presented Continual Visual Mapping (CVM), which uses a conceptual latent space created by a frozen Large Language Model (LLM) to train a small visual model. By learning to map the visual representations to a knowledgeable space, the visual model can learn new concepts based on the knowledge extracted through pre-learned textual information. We demonstrate the performance of CVM in class-incremental and domain-incremental learning using various benchmarks, achieving state-of-the-art results within resource-constrained environments. We also reveal scenarios where simple and small models can be more efficient (and surprisingly, even more effective) than methods based on large pre-trained models. In particular, these are scenarios where the low transferability of these methods affects the performance. Our learning strategy opens new research directions for CL methods to asynchronously take advantage of the knowledge of large pre-trained models efficiently and effectively mapping new knowledge into existing ones.

\section*{Acknowledgments}
Research partly funded by PNRR- PE00000013 - "FAIR - Future Artificial Intelligence Research" - Spoke 1 "Human-centered AI", funded by the European Commission under the NextGeneration EU programme, and by the Horizon EU project EMERGE (grant n. 101070918).

\bibliographystyle{unsrt}  
\bibliography{main}

\end{document}